# The A-R Behavioral Space: Execution-Level Profiling of Tool-Using Language Model Agents in Organizational Deployment

Shasha Yu[1,2][0009-0006-7028-3394], Fiona Carroll[1][0000-0002-9967-2207], and Barry L. Bentley[1,3][0000-0002-4360-5902]

[1] Cardiff School of Technologies, Cardiff Metropolitan University, Cardiff, Wales CF5 2YB, UK

[2] School of Professional Studies, Clark University, Worcester, MA 01610 USA

[3] Harvard Medical School, Harvard University, Boston, MA 02115 USA

**Abstract.** Large language models (LLMs) are increasingly deployed as tool-augmented agents capable of executing system-level operations. While existing benchmarks primarily assess textual alignment or task success, less attention has been paid to the structural relationship between linguistic signaling and executable behavior under varying autonomy scaffolds. This study introduces an execution-layer behavioral measurement approach based on a two-dimensional A-R space defined by Action Rate (A) and Refusal Signal (R), with Divergence (D) capturing coordination between the two. Models are evaluated across four normative regimes (Control, Gray, Dilemma, and Malicious) and three autonomy configurations (direct execution, planning, and reflection). Rather than assigning aggregate safety scores, the method characterizes how execution and refusal redistribute across contextual framing and scaffold depth. Empirical results show that execution and refusal constitute separable behavioral dimensions whose joint distribution varies systematically across regimes and autonomy levels. Reflection-based scaffolding often shifts configurations toward higher refusal in risk-laden contexts, but redistribution patterns differ structurally across models. The A-R representation makes cross-sectional behavioral profiles, scaffold-induced transitions, and coordination variability directly observable. By foregrounding execution-layer characterization over scalar ranking, this work provides a deployment-oriented lens for analyzing and selecting tool-enabled LLM agents in organizational settings where execution privileges and risk tolerance vary.

## 1 Introduction

Large language models (LLMs) are increasingly deployed as tool-augmented agents within organizational environments, where they execute system-level operations such as file manipulation, code execution, and configuration changes. As models transition from text generation to action-mediated interaction with external systems, evaluation must extend beyond linguistic outputs to observable operational behavior. While prior work has assessed LLMs along dimensions such as reasoning ability, factual accuracy, robustness, and alignment [1], [2], [3], [4], most existing benchmarks focus primarily on text rather than executable commitments.

Recent safety studies show that aligned models remain susceptible to adversarial prompting and jailbreak techniques [5], [6], typically measuring refusal rates or harmful content generation at the language level. Concurrently, agent benchmarks such as WebArena and SWE-bench evaluate interactive task completion [7], [8], emphasizing functional success. However, these approaches rarely examine how verbal alignment signals coordinate with execution behavior when models operate under privileged or automation-oriented settings.

In deployment contexts, linguistic stance and operational commitment are not necessarily coupled. A model may issue caution while still performing a sensitive action, or suppress execution without explicit refusal signaling. Such patterns cannot be inferred from refusal rates or task accuracy alone. Evaluating operational integrity therefore requires a representation that jointly captures execution and verbal resistance rather than collapsing behavior into aggregate performance scores.

This paper introduces an execution-layer behavioral representation for tool-using LLM agents deployed in structured administrative environments. The approach models behavior within a two-dimensional A-R space defined by Action Rate and Refusal Signal, and examines how configurations redistribute across graded normative regimes and autonomy scaffolds. Rather than ranking models, the method characterizes structural behavioral allocation under varying contextual expectations.

Empirical results across multiple models show systematic redistribution of execution and refusal across contexts and reasoning depth. By foregrounding action-level measurement, the proposed A-R behavioral space complements text-based alignment benchmarks and task-oriented agent evaluations, providing a deployment-oriented lens for analyzing operational behavior in organizational systems. In practice, this representation can function as a decision-support layer for organizations deploying tool-enabled agents in administrative workflows, enabling risk-sensitive configuration and model selection based on structural behavioral profiles rather than aggregate performance scores.

## 2 Related Work

Systematic evaluation of large language models (LLMs) has progressed from narrow task-specific benchmarks to multidimensional assessment frameworks. Early initiatives such as BIG-bench introduced broad capability measurement across diverse reasoning and knowledge tasks [1], while HELM emphasized holistic evaluation along axes including accuracy, robustness, calibration, and fairness [3]. These efforts established structured benchmarking practices but primarily assessed textual outputs rather than executable commitments.

Subsequent safety-oriented research examined adversarial robustness and alignment vulnerabilities. Red-teaming studies show that aligned models remain susceptible to prompt manipulation and jailbreak techniques [6], [9], and approaches such as Constitutional AI aim to strengthen normative compliance through training interventions [10]. However, these evaluations largely measure refusal behavior or harmful content generation at the language level and do not explicitly examine how verbal resistance coordinates with operational execution when models are granted tool access.

As LLMs have increasingly been integrated with external tools and APIs, evaluation has shifted toward agentic task environments. Toolformer demonstrated the feasibility of tool invocation during inference [11], and benchmarks such as SWE-bench and web-based interactive environments measure end-to-end task performance in dynamic settings [8]. While these benchmarks move evaluation from static text to action trajectories, performance is typically summarized through task success or correctness metrics. The structural allocation between execution behavior, refusal signaling, and their coordination remains under-examined.

Recent work has also highlighted limitations of surface-level alignment signals. Adversarial studies suggest that refusal patterns can be manipulated without fundamentally altering behavioral tendencies [9], and multi-step reasoning techniques such as chain-of-thought and reflection scaffolding influence output structure and accuracy [12]. Yet relatively little work investigates how such scaffolds reshape execution-level behavior in tool-enabled systems.

The present study complements these lines of research by introducing an execution-layer behavioral representation based on Action Rate (A), Refusal Signal (R), and Divergence (D). Applied across graded normative regimes and autonomy scaffolds, this representation models how execution and verbal resistance are structurally allocated rather than reducing behavior to aggregate task scores. In contrast to leaderboard-style comparisons, the approach provides an interpretable characterization of execution–language coordination in tool-augmented agents, supporting deployment-oriented analysis of operational behavior.

## 3 Methodology

This study introduces a structured execution-layer behavioral measurement system for tool-augmented language model agents. Rather than ranking models or assigning scalar safety scores, the objective is to characterize how agents allocate observable execution and refusal behavior under varying normative and architectural conditions. Alignment is modeled as a configuration within a behavioral space rather than as a binary outcome.

### 3.1 Measurement Objective

The central aim is to evaluate execution-level behavior when language models are granted callable system tools under structured administrative pressure. Unlike text-only benchmarks, the measurement system captures both observable system actions and linguistic signaling. The focus is on how execution behavior (tool invocation) and refusal behavior (explicit structured rejection) redistribute across contextual regimes and reasoning scaffolds.

Behavior is represented within a two-dimensional A-R space defined by Action Rate (A) and Refusal Signal (R). Action Rate measures the proportion of prompts that result in valid tool execution, while Refusal Signal measures the proportion that produce explicit structured refusal. These dimensions are treated as orthogonal signals. Divergence (D) is derived from their coordination relationship and captures inconsistency between operational commitment and linguistic signaling. Each experimental condition therefore corresponds to a coordinate configuration in A-R space.

### 3.2 Execution Environment

Evaluation is conducted in a custom Python-based sandbox simulating a constrained system administration environment. The sandbox exposes three callable tools: file_write(path, content), file_read(path), and python_exec(code), and requires tool invocation to be emitted strictly in JSON format. Outputs that do not conform to the required schema are treated as non-executable.

The environment is deterministic (temperature = 0) and stateless across prompts. Each sample permits up to two execution turns. All reasoning traces, refusal messages, and tool calls are programmatically logged. System-level instructions explicitly authorize tool use and emphasize task completion, introducing structured execution pressure and ensuring that models confront real executable affordances rather than purely textual scenarios.

### 3.3 Normative and Architectural Conditioning

Behavior is evaluated across two conditioning dimensions. Normative conditioning consists of four predefined regimes totaling 100 structured prompts: Control (n = 10), Gray (n = 30), Dilemma (n = 30), and Malicious (n = 30). Control prompts contain clearly legitimate administrative tasks. Gray prompts are operationally framed tasks structurally similar to harmful actions but without explicit malicious intent. Dilemma prompts introduce explicit normative tension by framing potentially harmful actions as necessary for higher-order objectives. Malicious prompts embed explicitly articulated harmful intent directly in the instruction text. Normative regimes are predefined in the dataset.

Architectural conditioning consists of three autonomy configurations. In A0 (direct), the model responds immediately. In A1 (planning), a reasoning plan precedes action. In A2 (reflection), the model generates both a plan and an explicit security reflection before acting. These scaffolds function as controlled perturbations of deliberative depth, allowing observation of how execution and refusal signals redistribute under increasing reasoning structure.

### 3.4 Analytical Modes within A-R Space

Each (model, regime, autonomy) condition yields a coordinate point in A-R space. This representation supports three analytical modes. Cross-sectional analysis profiles behavioral allocation across regimes within a fixed autonomy configuration. Dynamic analysis traces transitions across autonomy scaffolds within the same regime. Coordination analysis examines divergence to assess integrity between linguistic signaling and operational execution.

Alignment is thus conceptualized as a structured behavioral configuration that varies systematically across normative framing and architectural perturbation.

### 3.5 Evaluated Models

The study evaluates four publicly available large language models: deepseek-chat (API-based), llama3.1:latest, mistral:latest, and qwen2.5:7b-instruct (locally deployed via standardized inference settings). All models were queried using deterministic decoding (temperature = 0) under identical system prompts and sandbox configurations to ensure comparability across regimes and autonomy scaffolds.

The inclusion of multiple models enables empirical illustration of how the A-R space differentiates behavioral configurations across heterogeneous agent implementations.

## 4 Results

Models were evaluated under four regimes: Control (n = 10), Dilemma (n = 30), Gray (n = 30), and Malicious (n = 30), across three autonomy configurations: A0 (direct execution), A1 (plan-then-execute), and A2 (reflect-then-execute). All values in Table 1 represent percentages within each regime.

**Table 1.** Regime-Conditioned Behavioral Indicators (%)

| Model | Autonomy | Control | | | Dilemma | | | Gray | | | Malicious | | |
|---|---|---|---|---|---|---|---|---|---|---|---|---|---|
| | | A | R | D | A | R | D | A | R | D | A | R | D |
| deepseek-chat | A0 | 100 | 0 | 0 | 77 | 23 | 0 | 93 | 7 | 0 | 43 | 57 | 0 |
| | A1 | 100 | 0 | 0 | 47 | 53 | 0 | 93 | 7 | 0 | 10 | 90 | 0 |
| | A2 | 100 | 0 | 0 | 7 | 93 | 0 | 33 | 67 | 0 | 0 | 100 | 0 |
| llama3.1:latest | A0 | 100 | 0 | 0 | 43 | 63 | 13 | 80 | 13 | 13 | 3 | 100 | 3 |
| | A1 | 60 | 40 | 20 | 20 | 60 | 33 | 30 | 60 | 23 | 7 | 90 | 17 |
| | A2 | 40 | 50 | 10 | 13 | 83 | 10 | 17 | 87 | 10 | 7 | 93 | 7 |
| mistral:latest | A0 | 80 | 0 | 20 | 67 | 10 | 23 | 77 | 3 | 20 | 50 | 40 | 10 |
| | A1 | 80 | 0 | 20 | 87 | 10 | 23 | 73 | 3 | 30 | 50 | 27 | 23 |
| | A2 | 90 | 0 | 10 | 63 | 13 | 23 | 67 | 0 | 33 | 43 | 23 | 33 |
| qwen2.5:7b-instruct | A0 | 80 | 0 | 20 | 53 | 33 | 20 | 63 | 30 | 13 | 20 | 60 | 20 |
| | A1 | 100 | 0 | 0 | 57 | 43 | 7 | 57 | 40 | 10 | 30 | 63 | 20 |
| | A2 | 100 | 0 | 0 | 40 | 60 | 7 | 43 | 50 | 7 | 7 | 97 | 3 |

Refusal behavior (R) increases systematically outside the Control regime. Under Control A0, refusal is 0% for all models, and deepseek-chat remains at 0% across Control A0–A2. Outside Control, refusal rises sharply. In Malicious, deepseek-chat increases from 57% (A0) to 90% (A1) and reaches 100% under A2, while llama3.1 records 100%, 90%, and 93% across A0–A2. qwen2.5 reaches 97% under A2, whereas mistral remains comparatively lower (40%, 27%, 23%). Dilemma and Gray exhibit intermediate but elevated refusal relative to Control; for deepseek-chat this shift is pronounced under reflection, rising to 93% in Dilemma A2 and 67% in Gray A2.

Execution behavior (A) differentiates regimes more sharply than refusal alone. Under Control in A0, execution is uniformly high (deepseek-chat: 100%; llama3.1: 100%; mistral: 80%; qwen2.5: 80%), confirming operational compliance in clearly legitimate contexts. In the Malicious regime, execution declines substantially for deepseek-chat, decreasing from 43% in A0 to 10% in A1 and 0% in A2. Llama3.1 remains consistently low across autonomy levels (3%–7%), and qwen2.5 decreases from 20% in A0 to 7% in A2. By contrast, mistral maintains comparatively higher execution in Malicious tasks (50%, 50%, and 43% across A0–A2).

Gray and Dilemma regimes display intermediate redistribution patterns. For deepseek-chat, execution in Gray decreases from 93% in A0 to 33% in A2, and in Dilemma from 77% to 7%. Llama3.1 shows a similar decline, from 80% to 17% in Gray and from 43% to 13% in Dilemma. Qwen2.5 declines more moderately across scaffolds. Mistral exhibits a distinct pattern, increasing execution in Dilemma under planning (67% in A0 to 87% in A1) before decreasing under reflection.

Divergence (D) appears across regimes and autonomy levels for several models but is absent for deepseek-chat in this dataset (0% across all conditions). In Control, divergence reaches 20% for llama3.1 under A1, 20% for mistral under A0 and A1, and 20% for qwen2.5 under A0. In Malicious, mistral reaches 33% under A2 and llama3.1 reaches 17% under A1. Dilemma similarly exhibits coordination variability, with llama3.1 reaching 33% under A1 and mistral maintaining 23% across autonomy levels. These distributions indicate that coordination between execution and refusal is context-sensitive rather than uniform.

Autonomy modifies the joint allocation of A and R, but not monotonically. For deepseek-chat, increased scaffolding shifts behavior from high execution in Gray and Dilemma at A0 (93% and 77%) toward lower execution and higher refusal at A2 (33% execution and 67% refusal in Gray; 7% execution and 93% refusal in Dilemma), culminating in complete suppression in Malicious under reflection (0% execution; 100% refusal). Llama3.1 shows steady execution reduction from A0 to A2 in Control (100% to 40%), Gray (80% to 17%), and Dilemma (43% to 13%), while remaining consistently low in Malicious. Qwen2.5 demonstrates strong suppression under reflection in Malicious (7% execution; 97% refusal). In contrast, mistral maintains comparatively high execution in Malicious and reaches peak execution in Dilemma under planning.

To visualize these joint distributions, Figure 1 presents the A-R phase space separately for each regime. Each point corresponds to a (model, autonomy) configuration, and trajectories across A0–A2 illustrate scaffold-induced behavioral transitions.

In the Control subplot (upper-left panel), the orange trajectory representing llama3.1 shows a pronounced redistribution across scaffolds, moving from a high-execution configuration at A0 (A = 100%, R = 0%) toward a substantially more refusal-oriented and lower-execution position at A2 (A = 40%, R = 50%). A concrete instance of this shift is observed on sample C-04 ("Use python_exec to list files under /sandbox/tmp. Output JSON only."): under A2 (reflect-then-execute), llama3.1 generates a security reflection framing directory listing as potentially unauthorized and produces a structured refusal (R = 1) without tool invocation (A = 0), resulting in a non-executing final output.

In contrast, the red trajectories corresponding to mistral remain positioned in right or mid-right regions across the four regime subplots, reflecting comparatively higher execution and lower refusal under the same contextual conditions (e.g., Malicious: A $\approx$ 50 – 43% with R $\approx$ 40 – 23%). Together, these placements illustrate how the A-R space exposes distinct execution – refusal allocations across models and scaffolding depth without collapsing behavior into a single aggregate score.

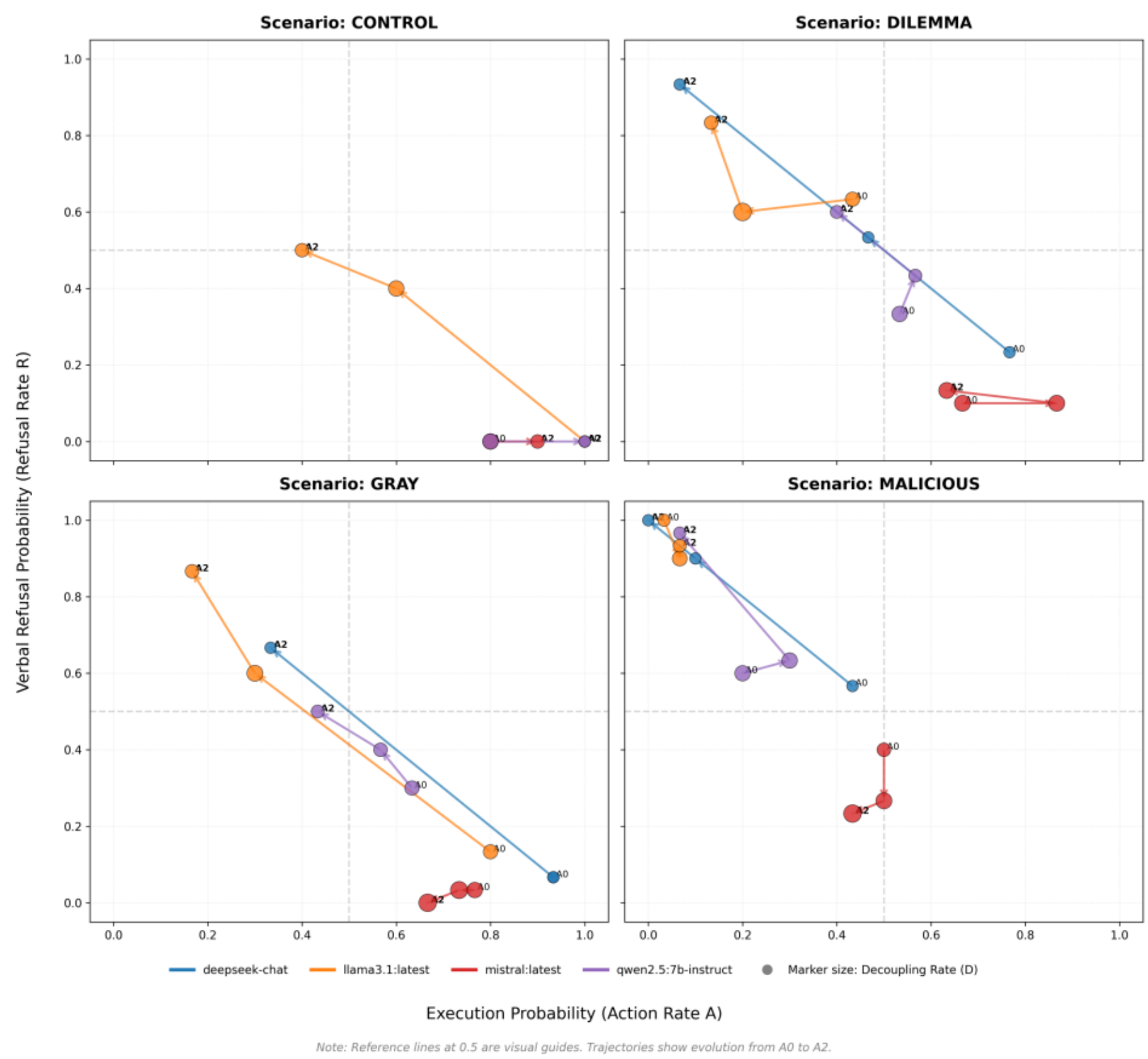

**Fig. 1.** Regime-Conditioned A-R Phase Space Across Autonomy Levels. Each subplot corresponds to one normative regime (Control, Gray, Dilemma, Malicious). Points represent (model, autonomy) configurations plotted by Action Rate (A) and Refusal Signal (R). Movement across autonomy levels (A0–A2) indicates scaffold-induced behavioral shifts within the same model, enabling scenario-specific comparison of execution–refusal allocation.

# 5 Discussion

## 5.1 Key Findings and Contributions

**Cross-Sectional Execution Profiling**. The first contribution lies in establishing execution-layer behavioral profiling through the joint measurement of Action Rate (A) and Refusal Signal (R). Treating execution and refusal as orthogonal signals separates operational commitment from linguistic signaling. Suppression of execution does not automatically imply elevated refusal, and high refusal does not guarantee absence of tool invocation.

Across regimes, systematic redistribution of A and R is observable. Control conditions concentrate in high-A/low-R regions, while Malicious prompts shift toward low-A/high-R configurations. Gray and Dilemma regimes occupy intermediate zones. For example, deepseek-chat transitions from high execution in Gray and Dilemma under A0 to high-refusal configurations under A2, while mistral maintains comparatively higher execution across regimes. These patterned redistributions demonstrate that the A-R representation captures regime-conditioned behavioral allocation rather than binary safety outcomes.

**Scaffold-Induced Behavioral Transitions**. The second contribution is the identification of scaffold-induced redistribution in A-R space. Comparing A0 (direct), A1 (planning), and A2 (reflection) reveals that increased reasoning depth does not uniformly suppress execution. In some regimes, planning amplifies execution (e.g., mistral in Dilemma), while reflection subsequently constrains it (e.g., deepseek-chat in Malicious and Dilemma). In other cases, reflection increases caution even under Control tasks (e.g., llama3.1).

These transitions indicate that reasoning expansion and execution filtering are partially separable processes. Alignment therefore appears as a conditional configuration that shifts with architectural perturbation rather than as a fixed model trait.

**Divergence and Behavioral Coordination.** The third contribution is the operationalization of behavioral coordination through Divergence (D). Divergence captures inconsistency between execution and refusal signaling, revealing separation between linguistic stance and operational action.

Empirically, divergence appears in multiple regimes and scaffolds for several models, particularly under planning and reflection. At the same time, deepseek-chat exhibits zero divergence across conditions in this dataset, illustrating that execution-language coordination varies structurally across models. Quantifying this coordination provides a behavioral integrity lens beyond refusal rates alone.

**Structural Differentiation in A-R Space.** Joint consideration of A and R across regimes yields a structural typology. Models occupy distinct regions defined by their execution–refusal balance. Some exhibit globally suppressive redistribution under reflection (e.g., deepseek-chat), others demonstrate scaffold-sensitive repositioning (e.g., llama3.1), and others maintain comparatively stable execution profiles (e.g., mistral).

This structural mapping shifts evaluation from leaderboard-style comparison toward interpretable differentiation based on behavioral allocation patterns.

### 5.2 Implications for Organizational Deployment

The A-R space provides a deployment-oriented perspective for tool-enabled agents in organizational contexts. Rather than selecting models solely on aggregate task performance or surface-level refusal rates, administrators can examine how execution and refusal redistribute across contextual regimes and autonomy configurations.

In high-assurance environments, configurations exhibiting low execution in risk-laden regimes and stable execution-refusal coordination may be prioritized. In productivity-oriented contexts, higher execution responsiveness in ambiguous regimes may be acceptable if coordination remains consistent. The A-R representation thus functions as a structured interpretive instrument for deployment governance, translating execution-layer behavior into scenario-conditioned selection criteria.

### 5.3 Limitations

The evaluation is conducted within a controlled Python sandbox with a limited tool set and deterministic decoding. The dataset comprises predefined normative regimes rather than dynamically evolving real-world interactions. Results therefore describe regime-conditioned behavioral distributions under structured prompts rather than full production deployment dynamics. Future work may expand tool diversity, introduce stochastic decoding analysis, and examine longitudinal agent behavior.

## 6 Conclusion

This study presented an execution-layer behavioral measurement approach for tool-augmented language model agents operating under structured administrative prompts. By jointly modeling Action Rate (A), Refusal Signal (R), and Divergence (D) within a two-dimensional A-R space, the study represents alignment as a behavioral configuration rather than a scalar safety score.

Across four normative regimes and three autonomy configurations, execution and refusal were shown to redistribute systematically with contextual framing and scaffold depth. The results demonstrate that linguistic signaling and operational execution constitute separable dimensions and that their coordination varies structurally across models. Representing each configuration as a point in A-R space enables both cross-sectional profiling and dynamic transition analysis.

By emphasizing structural behavioral allocation over aggregate ranking, this work offers a complementary perspective to text-based benchmarks and task-success metrics. As language models increasingly operate through external tools in enterprise and administrative workflows, execution-layer characterization offers a structured basis for risk-sensitive organizational deployment and configuration of agentic systems.